\documentclass[11pt,a4paper]{article}
\usepackage{authblk}
\usepackage[hyperref]{acl2021}
\usepackage{times}
\usepackage{multirow}
\usepackage{makecell}
\usepackage{latexsym}
\usepackage{booktabs}

\usepackage{microtype}

\aclfinalcopy 

\usepackage{latexsym}
\usepackage{amsmath}
\usepackage{amssymb}
\usepackage{algorithm}  
\usepackage{algorithmicx}  
\usepackage{algpseudocode}  
\usepackage{multicol} 
\usepackage{graphicx}
\usepackage{algorithm}  
\usepackage{algorithmicx}  

\usepackage{algpseudocode} 
\usepackage{subfigure,multirow}

\title{The USTC-NELSLIP Systems for Simultaneous Speech Translation Task at IWSLT 2021 }

\author[1,2]{Dan Liu}
\author[2]{Mengge Du}
\author[2]{Xiaoxi Li}
\author[1]{Yuchen Hu}
\author[1]{Lirong Dai }
\affil[1]{University of Science and Technology of China, Hefei, China}
\affil[2]{iFlytek Research, Hefei, China}

\affil[ ]{\textit{\{danliu, huyuchen\}@mail.ustc.edu.cn}}

\affil[ ]{\textit {lrdai@ustc.edu.cn}}

\affil[ ]{\textit{\{danliu, xxli16,mgdou\}@iflytek.com}}

\date{}

\begin{document}
\maketitle
\begin{abstract}
This paper describes USTC-NELSLIP's submissions to the IWSLT2021 Simultaneous Speech Translation task. We proposed a novel simultaneous translation model, Cross Attention Augmented Transducer (CAAT), which extends conventional RNN-T to sequence-to-sequence tasks without monotonic constraints, e.g., simultaneous translation. Experiments on speech-to-text (S2T) and text-to-text (T2T) simultaneous translation tasks shows CAAT achieves better quality-latency trade-offs compared to \textit{wait-k}, one of the previous state-of-the-art approaches.
Based on CAAT architecture and data augmentation, we build S2T and T2T simultaneous translation systems in this evaluation campaign.
Compared to last year's optimal systems, our S2T simultaneous translation system improves by an average of 11.3 BLEU for all latency regimes, and our T2T  simultaneous translation system improves by an average of 4.6 BLEU.
\end{abstract}

\section{Introduction}
This paper describes the submission to IWSLT 2021 Simultaneous Speech Translation task by National Engineering Laboratory for Speech and Language Information Processing (NELSLIP), University of Science and Technology of China, China.

Recent work in text-to-text simultaneous translation tends to fall into two categories,  fixed policy and flexible policy, represented by wait-k \citep{ma2018stacl} and monotonic attention \citep{arivazhagan-etal-2019-monotonic,ma2019monotonic} respectively. The drawback of fixed policy is that it may introduce over latency for some sentences and under latency for others.  Meanwhile, flexible policy often leads to difficulties in model optimization.

Inspired by RNN-T \citep{graves2012sequence}, we aim to optimize the marginal distribution of all expanded paths in simultaneous translation. However, we found it's impossible to calculate the marginal probability based on conventional Attention Encoder-Decoder \citep{sennrich2015neural} architectures (Transformer \citep{vaswani2017attention} included), which is due to the deep coupling between source contexts and target history contexts. To solve this problem, we propose a novel architecture, Cross Attention augmented Transducer (CAAT), and a latency loss function to ensure CAAT model works with an appropriate latency. In simultaneous translation, policy is integrated into translation model and learned jointly for CAAT model.

In this work, we build simultaneous translation systems for both text-to-text (T2T) and speech-to-text S2T) task. 
We propose a novel architecture, Cross Attention Augmented Transducer (CAAT), which significantly outperforms wait-k \citep{ma2018stacl} baseline in both text-to-text and speech-to-text simultaneous translation task. 
Besides, we adopt a variety of data augmentation methods, back-translation \citep{edunov2018understanding}, Self-training \citep{kim2016sequence} and speech synthesis with Tacotron2 \citep{shen2018natural}. Combining all of these and models ensembling, we achieved about 11.3 BLEU (in S2T task) and 4.6 BLEU (in T2T task) gains compared to the best performance last year. 


\section{Data}
\label{sec:data}
\subsection{Statistics and Preprocessing}
\paragraph{EN$\to$DE Speech Corpora}
The speech datasets used in our experiments are shown in Table~\ref{ta:dataset}, where MuST-C, Europarl and CoVoST2 are speech translation specific (speech, transcription and translation included), and LibriSpeech, TED-LIUM3 are speech recognition specific (only speech and transcription).
After augmented with speed and echo perturbation,
we use Kaldi \citep{povey2011kaldi} to extract 80 dimensional log-mel filter bank features, computed with a $25ms$ window size and a $10ms$ window shift, and specAugment \citep{park2019specaugment} were performed during training phase.

\begin{table}[htbp]
    \centering
    \begin{tabular}{ccc}
        \toprule
        \textbf{Corpus} &  \textbf{Segments} & \textbf{Duration(h)} \\
        \midrule
        MuST-C & 250.9k & 448  \\
        Europarl &69.5k &155\\
        CoVoST2 & 854.4k &1090 \\
        LibriSpeech & 281.2k & 960\\
        TED-LIUM3 &268.2k &452 \\
        \bottomrule
    \end{tabular}
    \caption{Statistics of speech corpora. }
    \label{ta:dataset}
\end{table}

\paragraph{Text Translation Corpora}
\label{sec:datatext}
The bilingual parallel datasets for Englith to German(EN$\to$DE)  and English to Japanese (EN$\to$JA)  used are shown in Table~\ref{ta:dataset-mt}, and the monolingual datasets in English, German and Japanese are shown in Table~\ref{ta:dataset-monolingual}. And we found the Paracrawl dataset in EN$\to$DE task is too big to our model training, we randomly select a subset of 14M sentences from it.

\begin{table}[htbp]
    \centering
    \begin{tabular}{ccc}
        \toprule
       & \textbf{Corpus} &  \textbf{Sentences} \\
        \midrule
        \multirow{8}{*}{\textbf{EN$\to$DE}} & MuST-C(v2) & 229.7k \\
        &Europarl & 1828.5k \\
        &Rapid-2019 & 1531.3k \\
        &WIT3-TED & 209.5k \\
        &Commoncrawl & 2399.1k \\
        &WikiMatrix & 6227.2k \\
        &Wikititles & 1382.6k \\
        &Paracrawl & 82638.2k \\
        \midrule
        \multirow{6}{*}{\textbf{EN$\to$JA}} &WIT3-TED & 225.0k \\
        &JESC & 2797.4k \\
        &kftt & 440.3k \\
        &WikiMatrix & 3896.0k \\
        &Wikititles &706.0k \\
        &Paracrawl & 10120.0k \\
        \bottomrule
    \end{tabular}
    \caption{Statistics of text parallel datasets. }
    \label{ta:dataset-mt}
\end{table}

\begin{table}[htbp]
    \begin{tabular}{ccc}
        \toprule
        \textbf{Language} & \textbf{Corpus} & \textbf{Sentences} \\
        \midrule
        \multirow{2}{*}{EN}&  Europarl-v10 & 2295.0k \\
            & News-crawl-2019 & 33600.8k \\
        \midrule
        \multirow{2}{*}{DE}& Europarl-v10 & 2108.0k \\
            & News-crawl-2020 & 53674.4k \\
        \midrule
        \multirow{2}{*}{JA}& News-crawl-2019 & 3446.4k \\
            & News-crawl-2020 & 10943.3k \\
        \bottomrule
    \end{tabular}
    \caption{Statistics of monolingual datasets.  }
    \label{ta:dataset-monolingual}
\end{table}

For EN$\to$DE task, we directly use SentencePiece \citep{kudo2018sentencepiece} to generate a unigram vocabulary of size 32,000 for source and target language jointly. 
And for EN$\to$JA task, sentences in Japanese are firstly participled by MeCab \citep{kudo2006mecab}, and then a unigram vocabulary of size 32,000 is generated for source and target jointly similar to EN$\to$DE task.

During data preprocessing, the bilingual datasets are firstly filtered by length less than 1024 and length ratio of target to source $0.25<r<4$. In the second step, with a baseline Transformer model trained with only bilingual data, we filtered the mismatched parallel pairs with log-likelihood from the baseline model, threshold is set to $-4.0$ for EN$\to$DE task and $-5.0$ for EN$\to$JA task.
At last we keep 27.3 million sentence pairs for EN-DE task and 17.0 sentence pairs for EN$\to$JA task.

\subsection{Data Augmentation}
For text-to-text machine translation, augmented data from monolingual corpora in source and target language are generated by self-training \citep{he2019revisiting} and back translation \citep{edunov2018understanding} respectively. Statistics of the augmented training data are shown in Table~\ref{data:all}.
\begin{table}[htbp]
    \centering
    \begin{tabular}{lcc}
        \toprule
        \textbf{Data} & \textbf{EN$\to$DE} & \textbf{EN$\to$JA}  \\
        \midrule 
        Bilingual data & 27.3M  & 17.0M  \\
        \ \ +back-translation& 34.3M & 22.0M  \\
        \ \ \ \ +self-training& 41.3M & 27.0M\\
        \bottomrule
    \end{tabular}
    \caption{Augmented training data for text-to-text translation.}
    \label{data:all}
    \vspace{-1.5em}
\end{table}

We further extend these two data augmentation methods to speech-to-text translation, detailed as:
\begin{enumerate}
    \item Self-training: Maybe similar to sequence-level distillation \citep{kim2016sequence,ren2020simulspeech,liu2019endtoend}. Transcriptions of all speech datasets (both speech recognition and speech translation specific) are sent to a text translation model to generate text $y^{'}$ in target language, the generated $y^{'}$ with its corresponding speech are directly added to speech translation dataset. 
    \item Speech Synthesis:
    We employ Tacotron2  \citep{shen2018natural} with slightly modified by introducing speaker representations to both encoder and decoder as our text-to-speech (TTS) model architecture, and trained on MuST-C(v2) speech corpora to generate filter-bank speech representations.
    We randomly select 4M sentence pairs from EN$\to$DE text translation corpora and generate audio feature by speech synthesis.
     The generated filter bank features and their corresponding target language text are used to expand our speech translation dataset.
\end{enumerate}
    The expanded training data are shown in Table~\ref{data:speech_aug}. Besides, during the training period for all the speech translation tasks, we sample the speech data from the whole corpora with fixed ratio and the concrete ratio for different dataset is shown in Table~\ref{data:dataset ratio}.
    
    \begin{table}[htbp]
        \centering
        \begin{tabular}{lcc}
            \toprule
           \textbf{ Dataset} & \textbf{Segements} & \textbf{Duration(h)}\\
            \midrule 
            Raw S2T dataset & 1.17M & 1693  \\
            \ \ +self-training& 2.90M & 4799  \\
            \quad+Speech synthesis& 7.22M & 10424\\
            \bottomrule
        \end{tabular}
        \caption{Expanded speech translation dataset with self-training and speech synthesis.}
        \label{data:speech_aug}
        \vspace{-0.5em}
    \end{table}

    \begin{table}[htbp]
        \centering
        \begin{tabular}{ccc}
            \toprule
            \textbf{Dataset} & \textbf{Sample Ratio} \\
            \midrule 
            MuST-C & 2  \\
            Europarl & 1\\
            CoVoST2 & 1 \\
            LibriSpeech & 1\\
            TED-LIUM3 &2\\
            Speech synthesis & 5 \\
            \bottomrule
        \end{tabular}
        \caption{Sample ratio for different datasets during training period.}
        \label{data:dataset ratio}
        \vspace{-0.5em}
    \end{table}
    
\section{Methods and Models}
\label{sec:methods}
\subsection{Cross Attention Augmented Transducer}
\label{sec:CAAT}
\def \x {\mathbf{x}}
\def \y {\mathbf{y}}
\def \h {\mathbf{h}}
\def \H {H}

Let $\x$ and $\y$ denote the source and target sequence, respectively. The policy of simultaneous translation is denoted as an action sequence $\mathbf{p} \in \{R,W\}^{|\x|+|\y|}$ where $R$ denotes the READ action and $W$ the WRITE action. Another representation of policy is extending target sequence $\y$ to length $|\x|+|\y|$ with blank symbol $\phi$ as $\hat{y} \in \left(\mathbf{v}\cup \{\phi\}\right)^{|\x|+|\y|}$, where $\mathbf{v}$ is the vocabulary of the target language. The mapping from $\y$ to sets of all possible expansion $\hat{y}$ denotes as $\H(\x,\y)$.

Inspired by RNN-T \citep{graves2012sequence}, the loss function for simultaneous translation can be defined as the marginal conditional probability and expectation of latency metric
through all possible expanded paths:

\begin{equation}
    \begin{split}
    &\mathcal{L}(x,y) =\mathcal{L}_{nll}(x,y)+\mathcal{L}_{latency}(x,y) \\
    &=-\log{\sum_{\hat{y}}{p(\hat{y}|x)}} + \mathbb{E}_{\hat{y}}l(\hat{y})\\
    &= -\log\sum_{\hat{y}}p(\hat{y}|x) + \sum_{\hat{y}}{\Pr(\hat{y}|y,x)l(\hat{y})}
    \end{split}
    \label{eq:cat}	
\end{equation}

\begin{figure}[ht]
    \centering
    \includegraphics[scale=0.5]{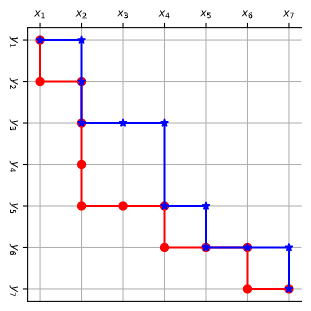}
    \caption{Expanded paths in simultaneous translation.}
    \label{fig:grid}
\end{figure}

Where $\Pr(\hat{y}|y,x)=\frac{p(\hat{y}|x)}{\sum_{\hat{y}^{'} \in \H(x,y)}p(\hat{y}^{'}|x)}$, and $\hat{y} \in \H(x,y)$ is an expansion of target sequence $\y$, and $l(\hat{y}$ is the latency of expanded path $\hat{y}$.

\begin{figure}[htbp]
    \centering
    \includegraphics[width=0.48\textwidth]{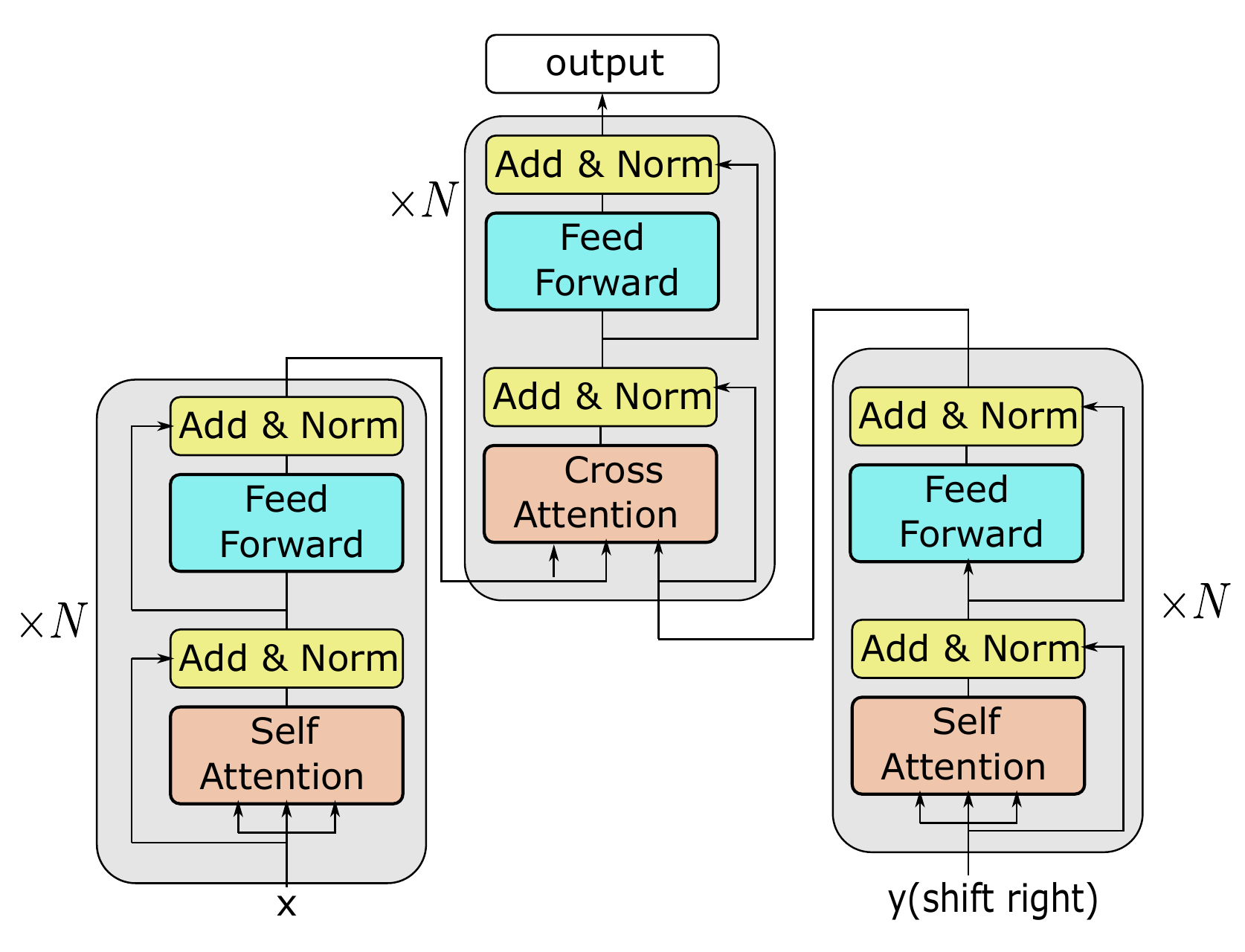}
    \caption{Architecture of CAAT based on Transformer.}
    \label{fig:caat}
\end{figure}

However, RNN-T is trained and inferenced based on source-target monotonic constraint, which means it isn't suitable for translation task. And the calculation of marginal probability $\sum_{\hat{y}\in \H(x,y)}\Pr(\hat{y}|x)$ is impossible for Attention Encoder-Decoder framework due to deep coupling of source and previous target representation.
As shown in Figure~\ref{fig:grid}, the decoder hidden states for the red path $\hat{y}^1$ and the blue path $\hat{y}^2$ is not equal at the intersection $s_2^1\ne s_2^2$. To solve this, we separate the source attention mechanism from the target history representation, which is similar to joiner and predictor in RNN-T. The novel architecture can be viewed as a extension version of RNN-T with attention mechanism augmented joiner, and is named as Cross Attention Augmented Transducer (CAAT). Figure~\ref{fig:caat} is the implementation of RAAT based on Transformer.

Computation cost of joiner in CAAT is significantly more expensive than that of RNN-T. The complexity of joiner is $\mathcal{O}(|\x|\cdot|\y|)$ during training, which means $\mathcal{O}(|\x|)$ times higher than conventional Transformer. We solve this problem by making decisions with decision step size $d>1$, and reduce the complexity of joiner from $\mathcal{O}(|\x|\cdot|\y|)$ to $\frac{\mathcal{O}(|\x|\cdot|\y|)}{d}$. Besides, to further reduce video memory consumption, we split hidden states into small pieces before sent into joiner, and recombine it for back-propagation during training.

As the latency loss is defined as marginal expectation over all expanded paths $\hat{y}$, \emph{mergeable} is also a requirement to the latency loss definition, which means latency loss through path $\hat{y}$ may be defined as $l(\hat{y}) = \sum_{k=1}^{|\x|+|\y|}l(\hat{y}_k)$ and $l(\hat{y}_k)$  is independent of $\hat{y}_{j^{'}\ne j}$. However, both Average Lagging \citep{ma2018stacl} and Differentiable Average Lagging \citep{arivazhagan-etal-2019-monotonic} do not meet this requirement. We hence introduce a novel latency function based on wait-0 as oracle latency as follows:
	\begin{equation}
		\begin{split}
			d(i,j) = \frac{1}{|\y|}\max\left(i-\frac{j\cdot|\x|}{|\y|},0\right) \\
			l(\hat{y}_k)=\begin{cases}
				0 & \text{if   } \hat{y}_k=\phi \\
				d(i_k,j_k) & \text{else}
			\end{cases}
		\end{split}
		\label{eq:lat_node}
	\end{equation}
	
	Where $i_k=\sum_{k^{'}=1}^{k}{I(\hat{y}_{k^{'}}= \phi)}$ and $j_k=\sum_{k^{'}=1}^{k}{I(\hat{y}_{k^{'}}\ne \phi)}$ denote READ and WRITE actions number before $\hat{y}_k$. The latency for the whole expanded path $\hat{y}$ can be defined as 
	\begin{equation}
		l(\hat{y}) = \sum_{k=1}^{|\hat{\y}|}l(\hat{y}_k)
		\label{eq:lat_loss}
	\end{equation}

Based on Eq.~\eqref{eq:lat_loss} the expectation of latency loss through all expanded paths may be defined as :
\begin{equation}
    \begin{split}
        \mathcal{L}_{latency}(x,y)&=\mathbb{E}_{\hat{y}\in H(x,y)}l(\hat{y}) \\
        &=\sum_{\hat{y}}{\Pr(\hat{y}|y,x)l(\hat{y})}
    \end{split}
    \label{eq:exp_lat}
\end{equation}
Latency loss and its gradients can be calculated by the forward-backward algorithm, similar to Sequence Criterion Training in ASR ~\citep{povey2005discriminative}.

At last, we add the cross entropy loss of offline translation model as an auxiliary loss to CAAT model training for two reasons. First we hope the CAAT model fall back to offline translation in the worst case; second, CAAT models is carried out in accordance with offline translation when source sentence ended. The final loss function for CAAT training is defined as follows:
	\begin{align}
		\mathcal{L}(x,y) &=\mathcal{L}_{CAAT}(x,y) +\lambda_{latency}\mathcal{L}_{latency}(x,y) \notag  \\
		& + \lambda_{CE}\mathcal{L}_{CE}(x,y) \notag \\
		&=-\log{\sum_{\hat{y}}{p(\hat{y}|x)}} \notag\\
		&+ \lambda_{latency}\sum_{\hat{y}}{\Pr(\hat{y}|y,x)d(\hat{y})} \notag \\
		&- \lambda_{CE}\sum_j\log{p(y_j|x,y_{<j})}  	\label{eq:loss_total}	
	\end{align}
	Where $\lambda_{latency}$ and $\lambda_{CE}$ are scaling factors corresponding to the $\mathcal{L}_{latency}$ and $\mathcal{L}_{CE}$. And we set $\lambda_1=\lambda_2=1.0$ if not specified.

\subsection{Streaming Encoder}
\label{sec:streamenc}
Unidirectional Transformer encoder \citep{arivazhagan-etal-2019-monotonic, ma2019monotonic} is not effective for speech data processing, because of the closely related to right context for speech frame $x_i$. Block processing \citep{dong2019self,wu2020streaming} is introduced for online ASR, but they lacks directly observing to infinite left context.

We process streaming encoder for speech data by block processing with right context and infinite left context.
First, input representations $\h$ is divided into overlapped blocks with block step $m$ and block size $m+r$. Each block consists of two parts, the main context $\mathbf{m}_n=\left[h_{m*n+1},\cdots,h_{(m+1)*n}\right]$ and the right context $\mathbf{r}_n=\left[h_{(m+1)*n},\cdots,h_{(m+1)*n+r}\right]$. The query, key and value of block $\mathbf{b}_n$ in self-attention can be described as follows:
\begin{align}
    \mathbf{Q}&=\mathbf{W}_q\left[\mathbf{m}_n, \mathbf{r}_n\right] \\
    \mathbf{K} &= \mathbf{W}_k\left[\mathbf{m}_1, \cdots,\mathbf{m}_n, \mathbf{r}_n\right] \\
    \mathbf{V} &= \mathbf{W}_v\left[\mathbf{m}_1, \cdots,\mathbf{m}_n, \mathbf{r}_n\right] 
\end{align}

By reorganizing input sequence and designed self-attention mask, training is effective by reusing conventional transformer encoder layers.
And unidirectional transformer can be regarded as a special case of our method with $\{m=1,r=0\}$. 
Note that the look-ahead window size in our method is fixed, which ensures increasing transformer layers won't affect latency.

\subsection{Text-to-Text Simultaneous Translation}
\label{sec:text_to_text}
We implemented both CAAT in Sec.~\ref{sec:CAAT} and wait-k \citep{ma2018stacl} systems for text-to-text simultaneous translation, both of them are implemented based on fairseq~\citep{ott2019fairseq}. 

All of wait-k experiments use the parameter settings based on big transformer ~\citep{vaswani2017attention} with unidirectional encoders, which corresponds to a 12-layer encoder and 6-layer decoder transformer with a embedding size of 1024, a feed forward network size of 4096, and 16 heads attention. 

Hyper-parameters of our CAAT model architectures are shown in Table~\ref{tab:hyp}. CAAT training requires significantly more GPU memory than conventional Transformer \citep{vaswani2017attention}, for the $\mathcal{O}\left(\frac{|x|\cdot|y|}{d}\right)$ complexity of joiner module. We mitigate this problem by reducing joiner hidden dimension for lower decision step size $d$.




\begin{table*}[htbp]
  \centering

    \begin{tabular}{ccccc}
    \toprule
    & \textbf{Parameters} & \textbf{S2T config} & \textbf{T2T config-A} &  \textbf{T2T config-B} \\
    \midrule
    \multirow{4}{*}{\textbf{Encoder}} & layers & 24 & 12 & 12 \\
    & attention heads & 8 & 16 &16\\
    & FFN dimension &2048 &4096 &4096 \\
    & embedding size & 512 & 1024 & 1024 \\
    \midrule
    \multirow{5}{*}{\textbf{Predictor}} & attention heads & 8 & 16 &16\\
    & FFN dimension &2048 &4096 &4096 \\
    & embedding size & 512 & 1024 & 1024 \\
    & output dimension & 512 & 512 & 1024 \\
    \midrule
    \multirow{4}{*}{\textbf{Joiner}} & attention heads & 8 & 8 &16\\
    & FFN dimension &1024 &2048 &4096 \\
    & embedding size & 512 & 512 & 1024 \\
    \midrule
    \multirow{2}{*}{\textbf{/}} & decision step size & \{16,64\} & \{4,10,16,32\} & \{10,32\} \\
    & latency scaling factor & \{1.0,0.2\} & \{1.0,0.2\} & 0.2 \\
    \bottomrule
    \end{tabular}%
  \caption{Parameters of CAAT in T2T and end-to-end S2T simultaneous translation. Noted that both predictor and joiner have 6 layers for T2T and S2T tasks, and the additional two parameters for end-to-end 2T simultaneous translation, which is the main context and right context described in Sec.\ref{sec:streamenc}, are set $m=32$ and  $r=16$ . }
  \label{tab:hyp}%
\end{table*}%

\begin{table}[htbp]
    \centering
    \begin{tabular}{lccc}
        \toprule
        \textbf{Dataset} & \textbf{BLEU}  \\
        \midrule 
        Original speech corpora & 21.24 \\
        \quad+self-training& 28.21 \\
        \quad\quad+Speech systhesis&29.72\\
        \bottomrule
    \end{tabular}
    \caption{Performance of offline speech translation on MuST-C(v2) tst-COMMON with different datasets.}
    \label{bleu:dataaug}
\end{table}


\subsection{Speech-to-Text Simultaneous Translation}
\subsubsection{End-to-End Systems}
The main system of End-to-End Speech-to-Text simultaneous Translation is based on the aforementioned CAAT structure. 
For speech encoder, two 2D convolution blocks are introduced before the stacked 24 Transformer encoder layers. Each convolution block consists of a 3-by-3 convolution layer with 64 channels and stride size as 2, and a ReLU activation function. Input speech features are downsampled 4 times by convolution blocks and flattened to 1D sequence as input to transformer layers. Other hyper-parameters are shown in Table~\ref{tab:hyp}.
The latency-quality trade-off may be adjusted by varying the decision step size $d$ and the latency scaling factor $\lambda_{latency}$. We submitted systems with best performance in each latency region.

\subsubsection{Cascaded Systems}
\label{method:st_cas}
The cascaded system consists of two modules, simultaneous automatic speech recognition (ASR) and simultaneous text-to-text Machine Translation (MT). Both simultaneous ASR and MT system are built with CAAT proposed in Sec.~\ref{sec:CAAT}. And we found the cascaded systems outperforms end-to-end system in medium and high latency region.

\subsection{Unsegmented Data Processing }
To deal with unsegmented data, we segment the input text based on sentence ending marks for T2T track. For S2T task, input speech is simply segmented into utterances with duration of 20 seconds and each segmented piece is directly sent to our simultaneous translation systems to obtain the streaming results. We found an abnormally large average lagging ($AL$) on IWSLT tst2018 test set based on existed SimuEval toolkit\citep{ma2020simuleval} and segment strategy, so relevant results are not presented here. A more reasonable latency criterion may be needed for unsegmented data in the future.

\section{Experiments}
\label{sec:exp}


\subsection{Effectiveness of CAAT}
\begin{figure}[htbp]
    \centering
    \includegraphics[width=0.48\textwidth]{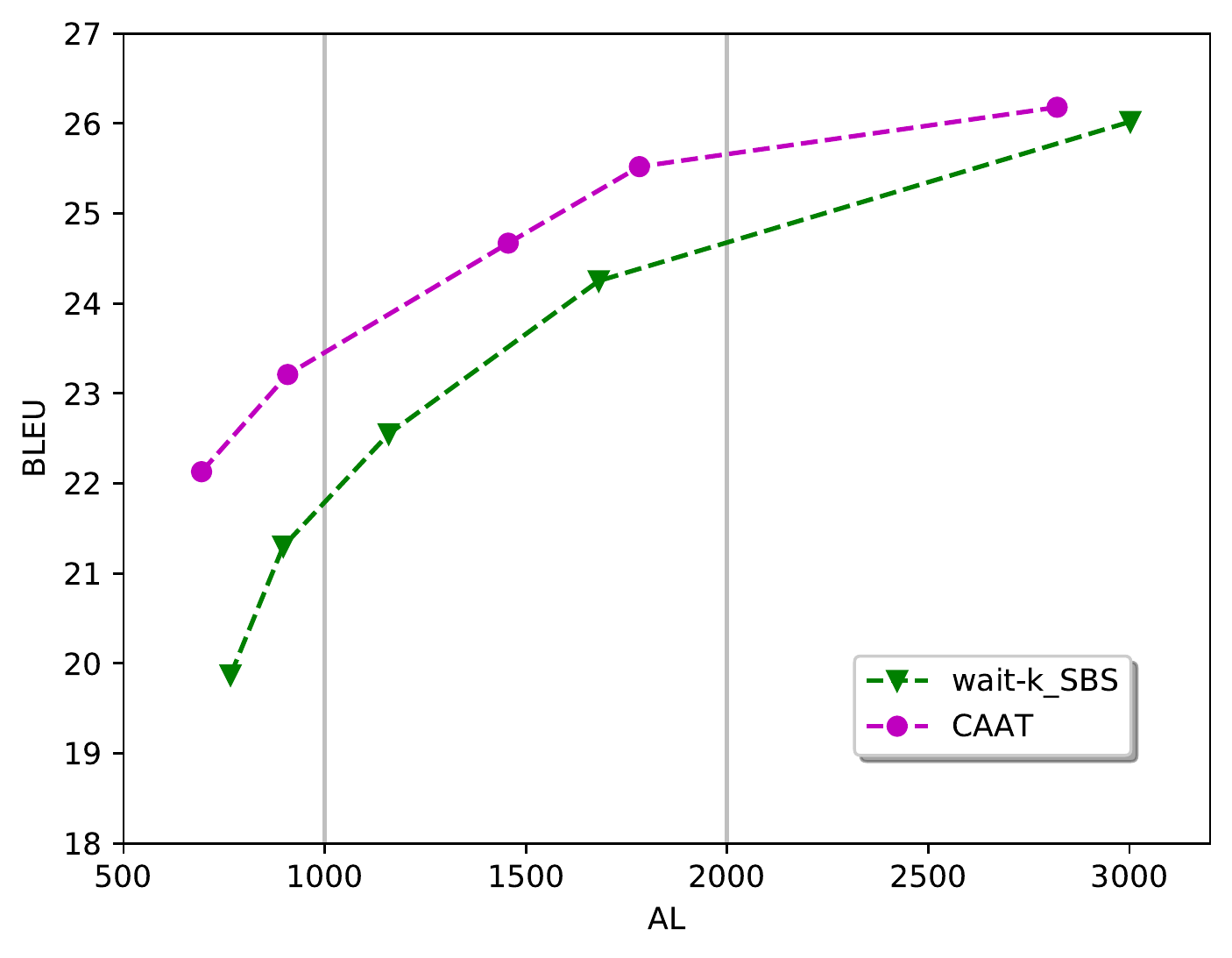}
    \caption{Comparison of CAAT and wait-k with SBS systems on EN$\to$DE Speech-to-Text simultaneous translation. }
    \label{fig:caat_albleu}
\end{figure}

To demonstrate the effectiveness of CAAT architecture, we compare it to wait-k with speculative beam search (SBS) \citep{ma2018stacl,zheng2019speculative}, one of the previous state-of-the-art.
The latency-quality trade-off curves on S2T and T2T tasks are shown in Figure~\ref{fig:caat_albleu} and Figure ~\ref{fig:t2tresult:a}. 
We can find that CAAT significantly outperforms wait-k with SBS, especially in low latency section($AL<1000ms$ for S2T track and $AL<3$ for T2T track). 

\subsection{Effectiveness of data augmentation}
In order to testify the effectiveness of data augmentation, we compare the results of different data augmentation methods based on the offline and simultaneous speech translation task. As demonstrated in Table~\ref{bleu:dataaug}, adding new generated target sentences into the training corpora by using Self-training gives a boost of nearly 7 BLEU points and speech synthesis provides the other 1.5 BLEU points increase on MuST-C(v2) tst-COMMON. As illustrated in Figure~\ref{fig:caat_albleu}, all the data augmentation methods are employed and provide nearly 3 BLEU points on average in the simultaneous task at different latency regimes. Note that our data augmentation methods alleviate the scarcity of parallel datasets in the End-to-End speech translation task and make a significant improvement.

    

\subsection{Text-to-Text Simultaneous Translation}
\begin{figure*}[htbp]
\subfigure[tst-COMMON(v2) on EN$\to$DE]{
    \label{fig:t2tresult:a}
    \begin{minipage}[t]{0.48\linewidth}
    \centering
    \includegraphics[width=1\textwidth]{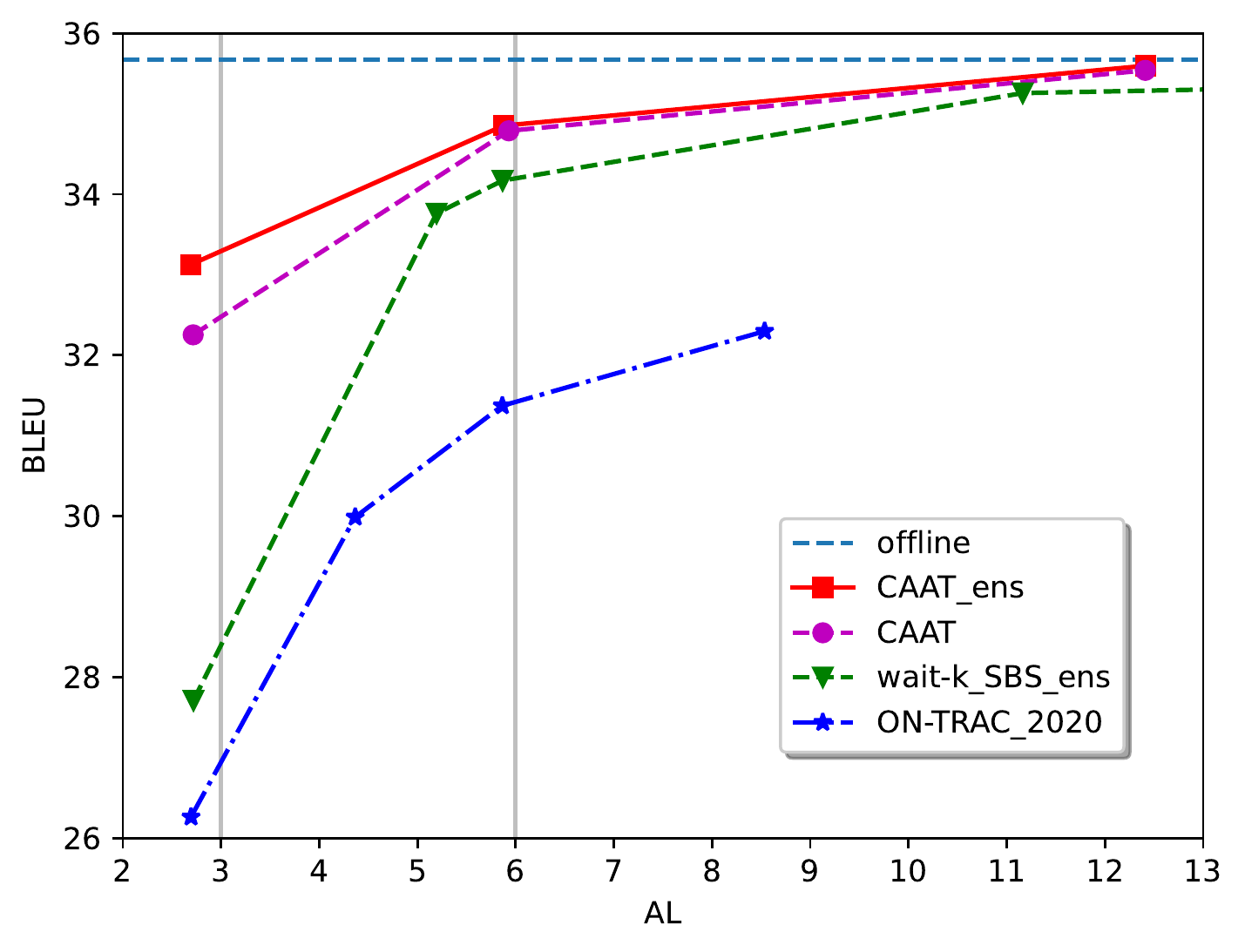}
    \end{minipage}%
}%
\subfigure[IWSLT dev2021 on EN$\to$JA]{
    \label{fig:t2tresult:b}
    \begin{minipage}[t]{0.48\linewidth}
    \centering
    \includegraphics[width=1\textwidth]{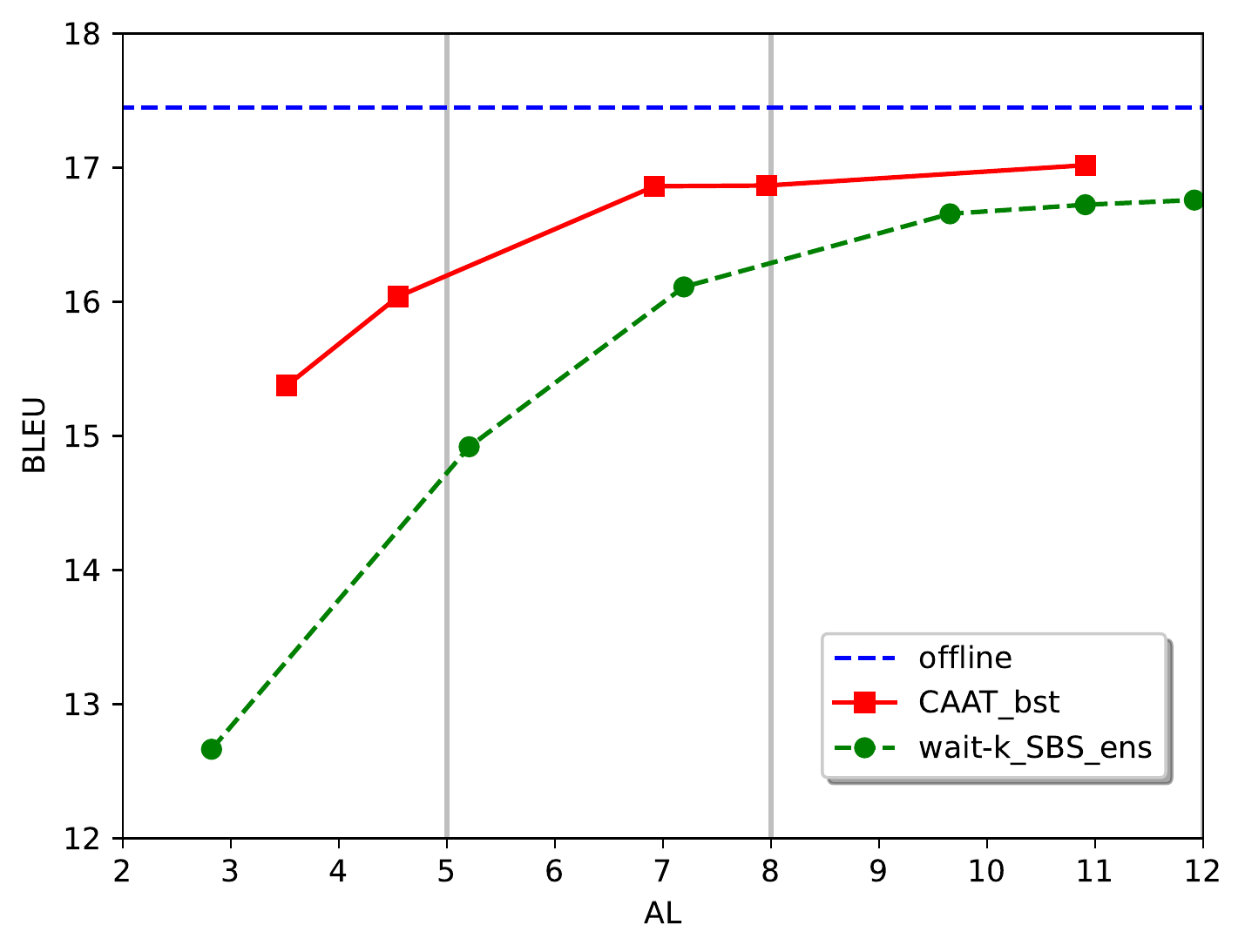}
    \end{minipage}
}
\caption{Latency-quality trade-offs of Text-to-Text simultaneous translation.}
\label{fig:t2tresult}
\end{figure*}

\paragraph{EN$\to$DE Task} The performances of text-to-text EN$\to$DE task is shown in Figure ~\ref{fig:t2tresult:a}. We can see that the performance of proposed CAAT is always better than that of wait-k with SBS and the best results from ON-TRAC in 2020 ~\citep{elbayad2020trac}, especially in low latency regime, and the performance of CAAT with model ensemble is nearly equivalent to offline result. Moreover, it can be further noticed from Figure ~\ref{fig:t2tresult:a} that the model ensemble can also improve the BLUE score more or less under different latency regimes, and the increase is quite obvious in low latency regime. Compared with the best result in 2020, we finally get improvement by 6.8 and 3.4 BLEU in low and high latency regime respectively.

\paragraph{En$\to$JA Task} Results of Text-to-Text simultaneous translation (EN$\to$JA) track are plotted in Figure ~\ref{fig:t2tresult:b}, where the curve naming CAAT\_bst is best performances in this track with or without model-ensembling method. Curves in this sub-figure show the similar conclusion to the former subsection, that the result of proposed CAAT significantly outperforms that of wait-k with SBS. While we can also find that the gap between CAAT and offline is more obvious (nearly 0.4 BLEU), this is mainly because parameters of joiner block for EN$\to$JA track in high-latency regime is reduced a lot from that for EN$\to$DE track, due to the unstable EN$\to$JA training.


\subsection{Speech-to-Text Simultaneous Translation}
\begin{figure}[htbp]
    \centering
    \includegraphics[width=0.48\textwidth]{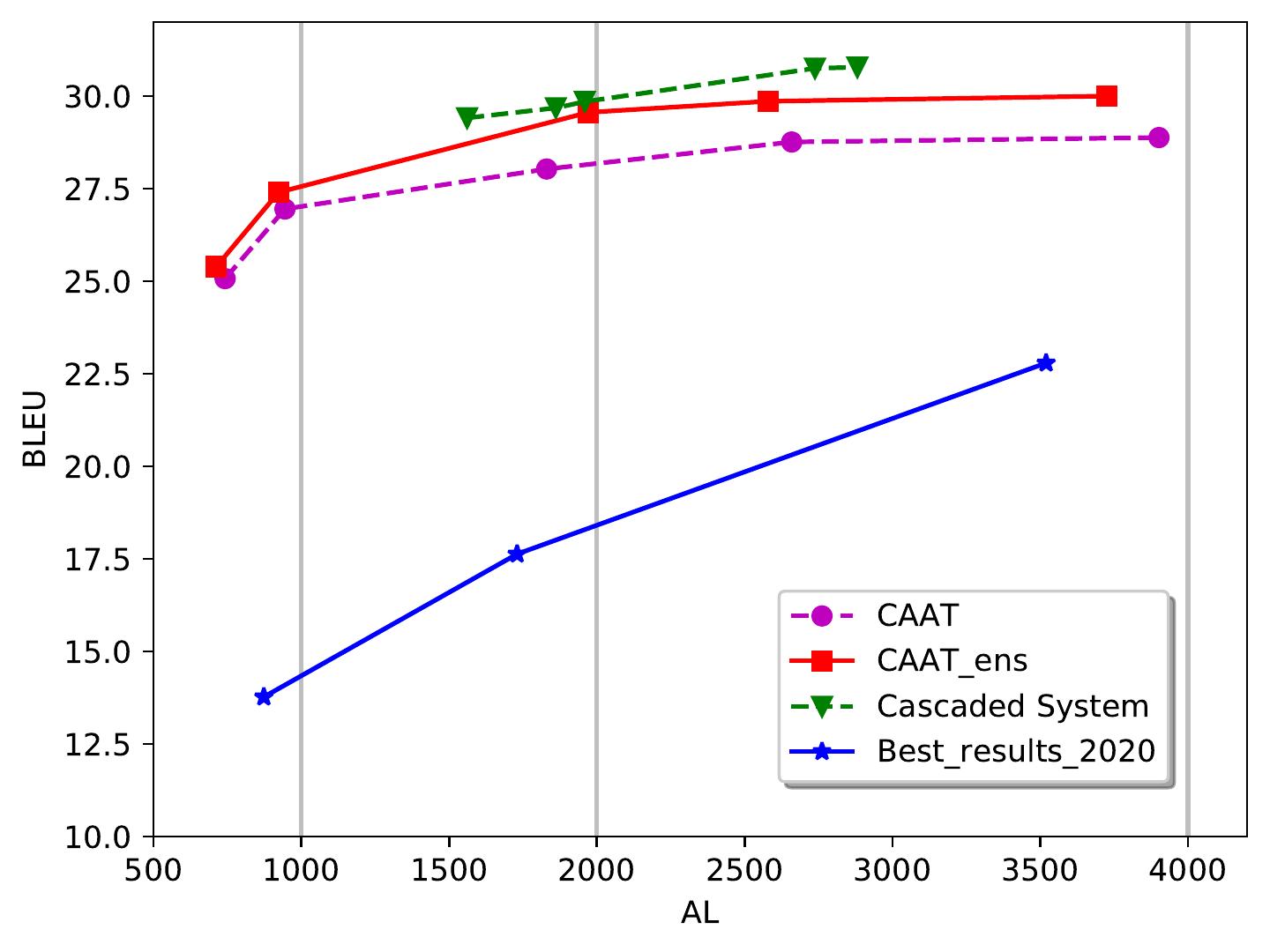}
    \caption{Latency-quality trade-offs of Speech-to-Text simultaneous translation on MuST-C(v2) tst-COMMON. }
    \label{fig:caat_final_res}
\end{figure}

\paragraph{End-to-End System} 
In this section, we discuss about our final results of End-to-End system based on CAAT. We tune the decision step size $d$ and latency scaling factor $\lambda_{latency}$ to meet different latency regime requirements. For low, medium and high latency, the corresponding $d$ and $\lambda_{latency}$ are set to (16,64,64) and (1.0,1.0,0.2) respectively.  We show our final latency-quality trade-offs in Figure~\ref{fig:caat_final_res}. Combined with our data augmentation methods and new CAAT model structure, it can be seen that our single model system has already outperformed the best results of last year in all latency regimes and provides 9.8 BLEU scores increase on average. Ensembling different models can further boost the BLEU scores by roughly 0.5-1.5 points at different latency regimes.

\paragraph{Cascaded System}
Under the cascaded setting, we paired two well-trained ASR and MT systems, where the WER of ASR system's performance is 6.30 with 1720.20 AL, and the MT system is followed by the config-A in Table ~\ref{tab:hyp}, whose results are 34.79 BLEU and 5.93 AL. We found the best medium and high-latency systems at decision step size pair $\left(d_{asr}, d_{mt}\right)$ with $(6, 10)$ and $(12, 10)$ respectively. Performance of cascaded systems are shown in Figure~\ref{fig:caat_final_res}. Note that under current configuration of ASR and MT systems, we can not provide valid results that satisfy the requirement of $AL$ at low latency regime since cascaded system usually has a larger latency compared to End-to End system. During the online decoding of the cascaded system, only after specific tokens are recognized by the ASR system, the translation model can further translate them to obtain the final result. The decoded results from ASR model first has a delay compared to the actual contents of the audio, and the two-steps decoding further accumulates the delay, which contributes to the higher latency compared to the End-to-End system. However, it still can be seen that cascaded system has significant advantages over End-to-End system at medium and high latency regime and it still has a long way to go for End-to-End system in the simultaneous speech translation task.

\section{Related Work}
\paragraph{Simultaneous Translation}
Recent work on simultaneous translation falls into two categories. The first category uses a fixed policy for the READ/WRITE actions and can thus be easily integrated into the training stage, as typified by \textit{wait-k} approaches \citep{ma2018stacl}.The second category includes models with a flexible policy learned and/or adaptive to current context, e.g., by Reinforcement Learning \citep{gu2016learning}, Supervise Learning \citep{zheng2019simultaneous} and so on. A special sub-category of flexible policy jointly optimizes policy and translation by monotonic attention customized to translation model, e.g., Monotonic Infinite Lookback (MILk) attention  \citep{arivazhagan-etal-2019-monotonic}  and Monotonic Multihead Attention (MMA) \citep{ma2019monotonic}.
We propose a novel method to optimize policy and translation model jointly, which is motivated by RNN-T \citep{graves2012sequence} in online ASR. Unlike RNN-T, the CAAT model removes the monotonic constraint, which is critical for considering reordering in machine translation tasks. The optimization of our latency loss is motivated by Sequence Discriminative Training in ASR  \citep{povey2005discriminative}. 

\paragraph{Data Augmentation}
As described in Sec.~\ref{sec:data}, the size of training data for speech translation is significantly smaller than that of text-to-text machine translation, which is the main bottleneck to improve the performance of speech translation. Self-training, or sequnece-level knowledge distillation by text-to-text machine translation model, is the most effective way to utilize the huge ASR training data \citep{liu2019endtoend,pino2020self}. On the other hand, synthesizing data by text-to-speech (TTS) has been demonstrated to be effective for low resource speech recognition task \citep{gokay2019improving,ren2019almost}. To the best of our knowledge, this is the first work to augment data by TTS for simultaneous speech-to-text translation tasks.

\section{Conclusion}
In this paper, we propose a novel simultaneous translation architecture, Cross Attention Augmented Transducer (CAAT), which significantly outperforms wait-k in both S2T and T2T simultaneous translation task. Based on CAAT architecture and data augmentation,  we build simultaneous translation systems on text-to-text and speech-to-text simultaneous translation tasks. We also build a cascaded speech-to-text simultaneous translation system for comparison. Both T2T and S2T systems 
achieve significant improvements over last year's best-performing systems.

\bibliographystyle{acl_natbib}
\bibliography{anthology,acl2021}


\end{document}